\documentclass[11pt,a4paper]{article}
\PassOptionsToPackage{breaklinks}{hyperref}
\usepackage[hyperref]{emnlp2019}
\usepackage{times}
\usepackage{latexsym}
\usepackage{amsmath}
\usepackage{mathrsfs}
\usepackage{amsfonts}
\usepackage{amssymb}
\usepackage{booktabs}
\usepackage{graphicx}
\usepackage{inconsolata}
\usepackage{listings}
\usepackage{url}
\usepackage[linesnumbered]{algorithm2e}
\SetKwComment{Comment}{$\triangleright$\ }{}

\aclfinalcopy 

\definecolor{javaGreen}{RGB}{63,127,95}

\lstdefinestyle{mysql}{
language=Sql,
stringstyle=\color{javaGreen},
deletekeywords={year}
}

\newcommand{\our}{Iyer-Simp}

\title{Learning Programmatic Idioms for Scalable Semantic Parsing}

\author{
Srinivasan Iyer$^\dagger$, Alvin Cheung$^\mathsection$ and Luke Zettlemoyer$^{\dagger\ddagger}$\\
$^\dagger$Paul G. Allen School of Computer Science and Engineering, Univ. of Washington, Seattle, WA \\
\tt{\{sviyer, lsz\}@cs.washington.edu} \\
$\mathsection$Department of Electrical Engineering and Computer Sciences, UC Berkeley, Berkeley, CA \\
\tt{akcheung@cs.berkeley.edu} \\
$\ddagger$ Facebook AI Research, Seattle\\
\tt{lsz@fb.com}\\
}\date{}

\begin{document}
\maketitle
\begin{abstract}
Programmers typically organize executable source code using high-level coding patterns or idiomatic structures such as nested loops, exception handlers and recursive blocks, rather than as individual code tokens. In contrast, state of the art (SOTA) semantic parsers still map natural language instructions to source code by building the code syntax tree one node at a time. In this paper, we introduce an iterative method to extract code idioms from large source code corpora by repeatedly collapsing most-frequent depth-2 subtrees of their syntax trees, and train semantic parsers to apply these idioms during decoding. Applying idiom-based decoding on a recent context-dependent semantic parsing task improves the SOTA by 2.2\% BLEU score while reducing training time by more than 50\%. This improved speed enables us to scale up the model by training on an extended training set that is 5$\times$ larger, to further move up the SOTA by an additional 2.3\% BLEU and 0.9\% exact match. Finally, idioms also significantly improve accuracy of semantic parsing to SQL on the ATIS-SQL dataset, when training data is limited.


\end{abstract}

\section{Introduction}

\begin{figure}[ht]
\includegraphics[width=0.9\linewidth]{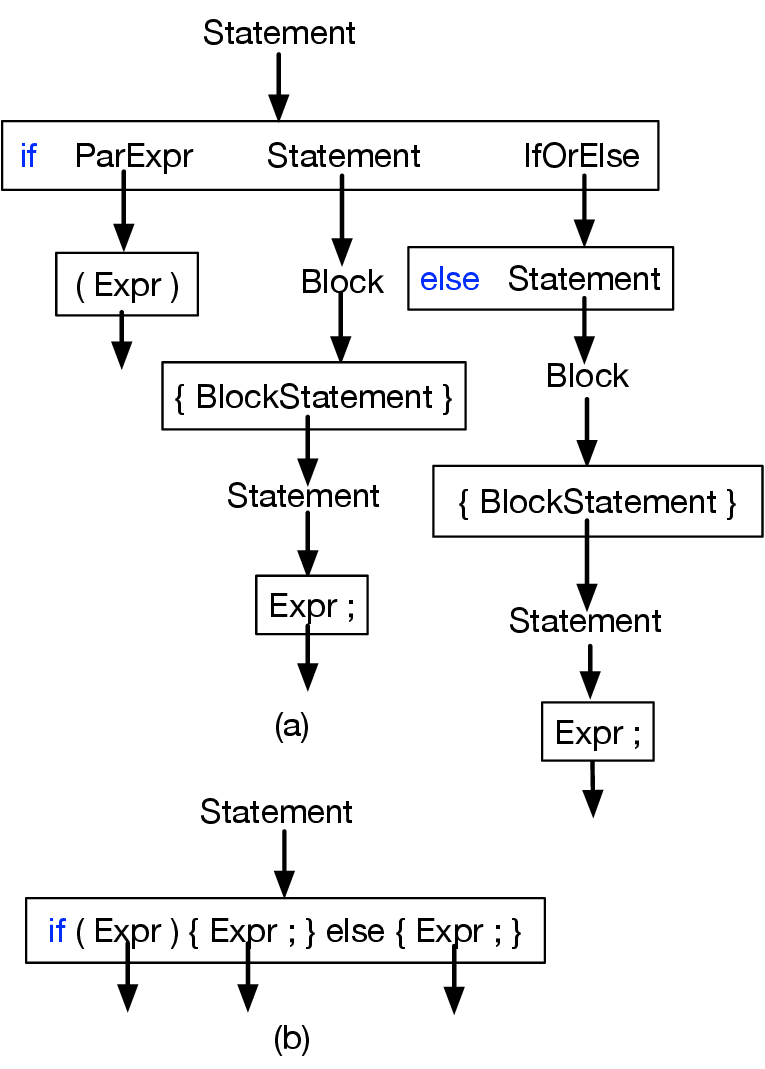}
\caption{(a) Syntax tree based decoding for semantic parsing uses as many as 11 rules (steps) to produce the outer structure of a very frequently used \texttt{if-then-else} code snippet. (b) Direct application of an \texttt{if-then-else} idiom during decoding leads to improved accuracy and training time.}
\label{fig:idiom_main}
\end{figure}

When programmers translate Natural Language (NL) specifications into executable source code, they typically start with a high-level plan of the major structures required, such as nested loops, conditionals, etc. and then proceed to fill in specific details into these components. We refer to these high-level structures (Figure \ref{fig:idiom_main} (b)) as code idioms \cite{allamanis2014mining}.  In this paper, we demonstrate how learning to use code idioms leads to an improvement in model accuracy and training time for the task of semantic parsing, i.e., mapping intents in NL into general purpose source code \cite{iyer-EtAl:2017:Long,ling2016}.

State-of-the-art semantic parsers are neural encoder-decoder models, where decoding is guided by the target programming language grammar \cite{yin-neubig:2017:Long,rabinovich-stern-klein:2017:Long,iyer2018} to ensure syntactically valid programs. For general purpose programming languages with large formal grammars, this can easily lead to long decoding paths even for short snippets of code. For example, Figure \ref{fig:idiom_main} shows an intermediate parse tree for a generic \texttt{if-then-else} code snippet, for which the decoder requires as many as eleven decoding steps before ultimately filling in the slots for the \texttt{if} condition, the \texttt{then} expression and the \texttt{else} expression. However, the \texttt{if-then-else} block can be seen as a higher level structure such as shown in Figure \ref{fig:idiom_main} (b) that can be applied in one decoding step and reused in many different programs. In this paper, we refer to frequently recurring subtrees of programmatic parse trees as {\em code idioms}, and we equip semantic parsers with the ability to learn and directly generate idiomatic structures as in Figure \ref{fig:idiom_main} (b).

We introduce a simple iterative method to extract idioms from a dataset of programs by repeatedly collapsing the most frequent depth-2 subtrees of syntax parse trees. Analogous to the byte pair encoding (BPE) method \cite{gage1994new,sennrich-haddow-birch:2016:P16-12} that creates new subtokens of words by repeatedly combining frequently occurring adjacent pairs of subtokens, our method takes a depth-2 syntax subtree and replaces it with a tree of depth-1 by removing all the internal nodes. This method is in contrast with the approach using probabilistic tree substitution grammars (pTSG) taken by \newcite{allamanis2014mining}, who use the \textit{explanation quality} of an idiom to prioritize idioms that are more interesting, with an end goal to suggest useful idioms to programmers using IDEs. Once idioms are extracted, we greedily apply them to semantic parsing training sets to provide supervision for learning to apply idioms. 

We evaluate our approach on two semantic parsing tasks that map NL into 1) general-purpose source code, and 2) executable SQL queries, respectively. On the first task, i.e., context dependent semantic parsing \cite{iyer2018} using the CONCODE dataset, we improve the state of the art (SOTA) by 2.2\% of BLEU score. Furthermore, generating source code using idioms results in a more than 50\% reduction in the number of decoding steps, which cuts down training time to less than half, from 27 to 13 hours. Taking advantage of this reduced training time, we further push the SOTA on CONCODE to an EM of 13.4 and a BLEU score of 28.9 by training on an extended version of the training set (with 5$\times$ the number of training examples). On the second task, i.e., mapping NL utterances into SQL queries for a flight information database (ATIS-SQL; \newcite{iyer-EtAl:2017:Long}), using idioms significantly improves denotational accuracy over SOTA models, when a limited amount of training data is used, and also marginally outperforms the SOTA when the full training set is used (more details in Section \ref{sec:results}).




\section{Related Work}

Neural encoder-decoder models have proved effective in mapping NL to logical forms \cite{dong2016} and also for directly producing general purpose programs \cite{iyer-EtAl:2017:Long,iyer2018}. \newcite{ling2016} use a sequence-to-sequence model with attention and a copy mechanism to generate source code. Instead of directly generating a sequence of code tokens, recent methods focus on constrained decoding mechanisms to generate syntactically correct output using a decoder that is either grammar-aware or has a dynamically determined modular structure paralleling the structure of the abstract syntax tree (AST) of the code \cite{rabinovich-stern-klein:2017:Long,krishnamurthy-dasigi-gardner:2017:EMNLP2017,yin-neubig:2017:Long}. \newcite{iyer2018} use a similar decoding approach but use a specialized context encoder for the task of context-dependent code generation. We augment these neural encoder-decoder models with the ability to decode in terms of frequently occurring higher level idiomatic structures to achieve gains in accuracy and training time.

Another different but related method to produce source code is using sketches, which are code snippets containing slots in the place of low-level information such as variable names, method arguments, and literals. \newcite{dong2018coarse} generate such sketches using programming language-specific sketch creation rules and use them as intermediate representations to train token-based seq2seq models that convert NL to logical forms. \newcite{hayati2018} retrieve sketches from a large training corpus and modify them for the current input; \newcite{murali2018neural} use a combination of neural learning and type-guided combinatorial search to convert existing sketches into executable programs, whereas \newcite{nye2019learning} additionally also generate the sketches before synthesising programs. Our idiom-based decoder learns to produce commonly used subtrees of programming syntax-trees in one decoding step, where the non-terminal leaves function as slots that can be subsequently expanded in a grammer-aware fashion. Code idioms can be roughly viewed as a tree-structured generalization of sketches, that can be automatically extracted from large code corpora for any programming language, and unlike sketches, can also be nested with other idioms or grammar rules.


 
 More closely related to the idioms that we use for decoding is \newcite{allamanis2014mining}, who develop a system (HAGGIS) to automatically mine idioms from large code bases. They focus on finding interesting and explainable idioms, e.g., those that can be included as preset code templates in programming IDEs. Instead, we learn frequently used idioms that can be easily associated with NL phrases in our dataset. The production of large subtrees in a single step directly translates to a large speedup in training and inference.
 
 Concurrent with our research, \newcite{shin2019program} also develop a system (PATOIS) for idiom-based semantic parsing and demonstrate its benefits on the Hearthstone \cite{ling2016} and Spider \cite{yu2018spider} datasets. While we extract idioms by collapsing frequently occurring depth-2 AST subtrees and apply them greedily during training, they use non-parametric Bayesian inference for idiom extraction and train neural models to either apply entire idioms or generate its full body.

\section{Idiom Aware Encoder-Decoder Models}

We aim to train semantic parsers having the ability to use idioms during code generation. To do this, we first extract frequently used idioms from the training set, and then provide them as supervision to the semantic parser's learning algorithm. 

Formally, if a semantic parser decoder is guided by a grammar $\mathscr{G} = (N, \Sigma, R)$, where $N$ and $\Sigma$ are the sets of non-terminals and terminals respectively, and $R$ is the set of production rules of the form $A \rightarrow \beta, A \in N, \beta \in \{N \cup \Sigma\}^*$, we would like to construct an idiom set $I$ with rules of the form $B \rightarrow \gamma, B \in N, \gamma \in \{N \cup \Sigma\}^*$, such that $B\overset{\geq 2}{\implies}\gamma$ under $\mathscr{G}$, i.e., $\gamma$ can be derived in two or more steps from $B$ under $\mathscr{G}$. For the example in Figure \ref{fig:idiom_main}, $R$ would contain rules for expanding each non-terminal, such as \textit{Statement $\rightarrow$ if ParExpr Statement IfOrElse} and \textit{ParExpr $\rightarrow$ \{ Expr \}}, whereas $I$ would contain the idiomatic rule \textit{Statement $\rightarrow$ if ( Expr ) \{ Expr ; \} else \{ Expr ; \} }. 

The decoder builds trees from $\hat{\mathscr{G}} = (N, \Sigma, R \cup I)$. Although the set of valid programs under both $\mathscr{G}$ and $\hat{\mathscr{G}}$ are exactly the same, this introduction of ambiguous rules into $\mathscr{G}$ in the form of idioms presents an opportunity to learn shorter derivations. In the next two sections, we describe the idiom extraction process, i.e., how $I$ is chosen, and the idiom application process, i.e., how the decoder is trained to learn to apply idioms.

\begin{algorithm}[h]
\SetKwProg{myproc}{Procedure}{}{end}

\myproc{Extract-Idioms($\mathscr{D}$, $\mathscr{G}$, $n$)}{
\KwIn{$\mathscr{D} \rightarrow$ Training Programs}
\KwIn{$\mathscr{G} \rightarrow$ Parsing Grammar}
\KwIn{$n \rightarrow$ Number of idioms }
  $T \gets \{\}$ \Comment{Stores all parse trees}
  \For{$d \in \mathscr{D}$}{
    $T \gets T$ $\cup $ Parse-Tree$(d, \mathscr{G})$ \\
  }
  
  $I \gets \{ \}$ \\
  \For{$i\gets1$ \KwTo $K$}{
    $s \gets $ Most-frequent Depth-2-Subtree$(T)$ \\
    \For{$t \in T$}{
      $t \gets $ Collapse-Subtree(t, s)
      }
    $I \gets I \cup \{s\}$
}
}
\myproc{Collapse-Subtree($t, s$)}{
\KwIn{$t \rightarrow $ parse-tree}
\KwIn{$s \rightarrow $ Depth 2 subtree}
\While{$s$ subtree-of $t$}{
  Replace $s$ in $t$ with Collapse$(s)$
}
}
\myproc{Collapse($s$)}{
\KwIn{$s \rightarrow $ Depth 2 subtree}
frontier $ \gets $ leaves$(s)$ \\
return Tree$(root(s), $frontier$)$ 
}
\caption{Idiom Extraction.}
\label{alg:extract}
\end{algorithm}

\section{Idiom Extraction}


Algorithm \ref{alg:extract} describes the procedure to add idiomatic rules, $I$, to the regular production rules, $R$. Our goal is to populate the set $I$ by identifying frequently occurring idioms (subtrees) from the programs in training set $\mathscr{D}$. Since enumerating all subtrees of every AST in the training set is infeasible, we observe that all subtrees $s'$ of a frequently occurring subtree $s$ are just as or more frequent than $s$, so we take a bottom-up approach by repeatedly collapsing the most frequent depth-2 subtrees. Intuitively, this can be viewed as a particular kind of generalization of the BPE \cite{gage1994new, sennrich-haddow-birch:2016:P16-12} algorithm for sequences, where new subtokens are created by repeatedly combining frequently occurring adjacent pairs of subtokens. Note that subtrees of parse trees have an additional constraint, i.e.,  either all or none of the children of non-terminal nodes are included, since a grammar rule has to be used entirely or not at all.


\begin{figure}[t]
\includegraphics[width=\linewidth]{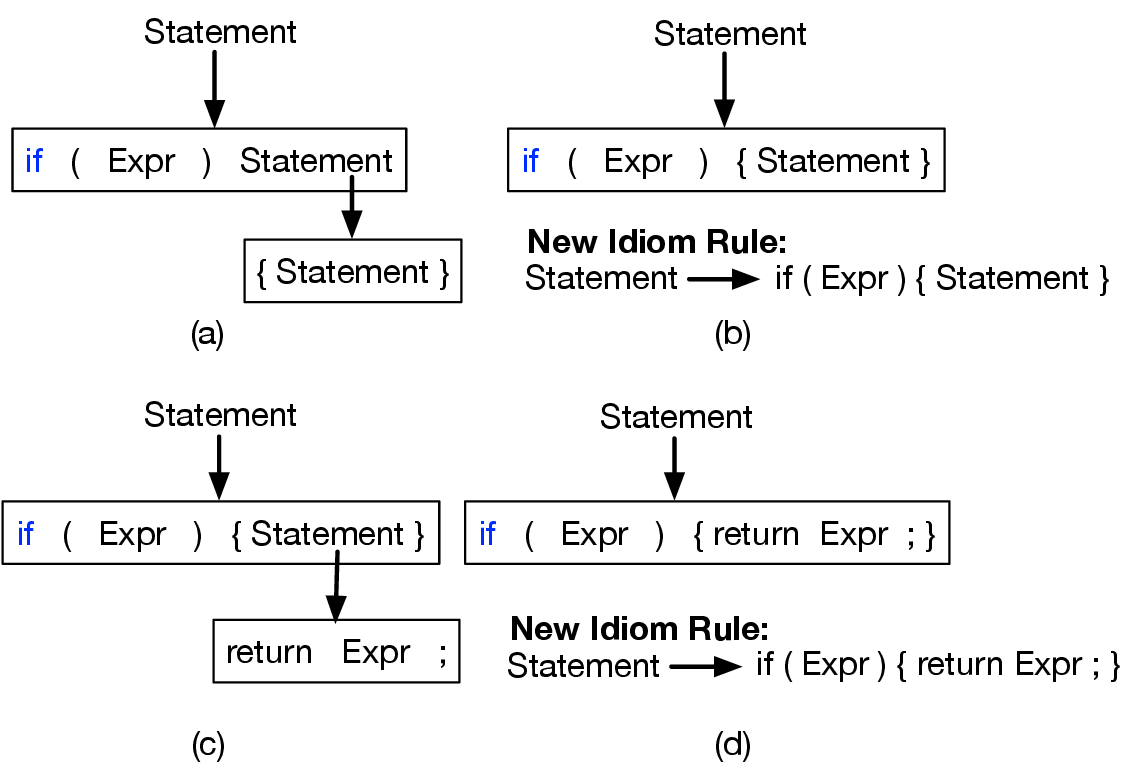}
\caption{Two steps of the idiom extraction process described in Algorithm \ref{alg:extract}. (a) First, we find the most frequent depth-2 syntax subtree under Grammar $\mathscr{G}$ in dataset $\mathscr{D}$, collapse it to produce a new production rule (b), and replace all occurrences in the dataset with the collapsed version. Next, (c) is the most frequent depth-2 subtree which now happens to contain (b) within it. (c) is collapsed to form an idiom (d), which is effectively a depth-3 idiom.}
\label{fig:idiom_extraction}
\end{figure}

We perform idiom extraction in an iterative fashion. We first populate $T$ with all parse trees of programs in $\mathscr{D}$ using grammar $\mathscr{G}$ (Step 4). Each iteration then comprises retrieving the most frequent depth-2 subtree $s$ from $T$ (Step 8), followed by post-processing $T$ to replace all occurrences of $s$ in $T$ with a collapsed (depth-1) version of $s$ (Step 10 and 17). The collapse function (Step 20) simply takes a subtree, removes all its internal nodes and attaches its leaves directly to its root (Step 22). The collapsed version of $s$ is a new idiomatic rule (a depth-1 subtree), which we add to our set of idioms, $I$ (Step 12). We illustrate two iterations of this algorithm in Figure \ref{fig:idiom_extraction} ((a)-(b) and (c)-(d)). Assuming (a) is the most frequent depth-2 subtree in the dataset, it is transformed into the idiomatic rule in (b). Larger idiomatic trees are learned by combining several depth-2 subtrees as the algorithm progresses. This is shown in Figure \ref{fig:idiom_extraction} (c) which contains the idiom extracted in (b) within it owing to the post-processing of the dataset after idiom (b) is extracted (Step 10 of Algorithm \ref{alg:extract}) which effectively makes the idiom in (d), a depth-3 idiom. We perform idiom extraction for $K$ iterations. In our experiments we vary the value of $K$ based on the number of idioms we would like to extract.

\begin{figure}[t]
\includegraphics[width=\linewidth]{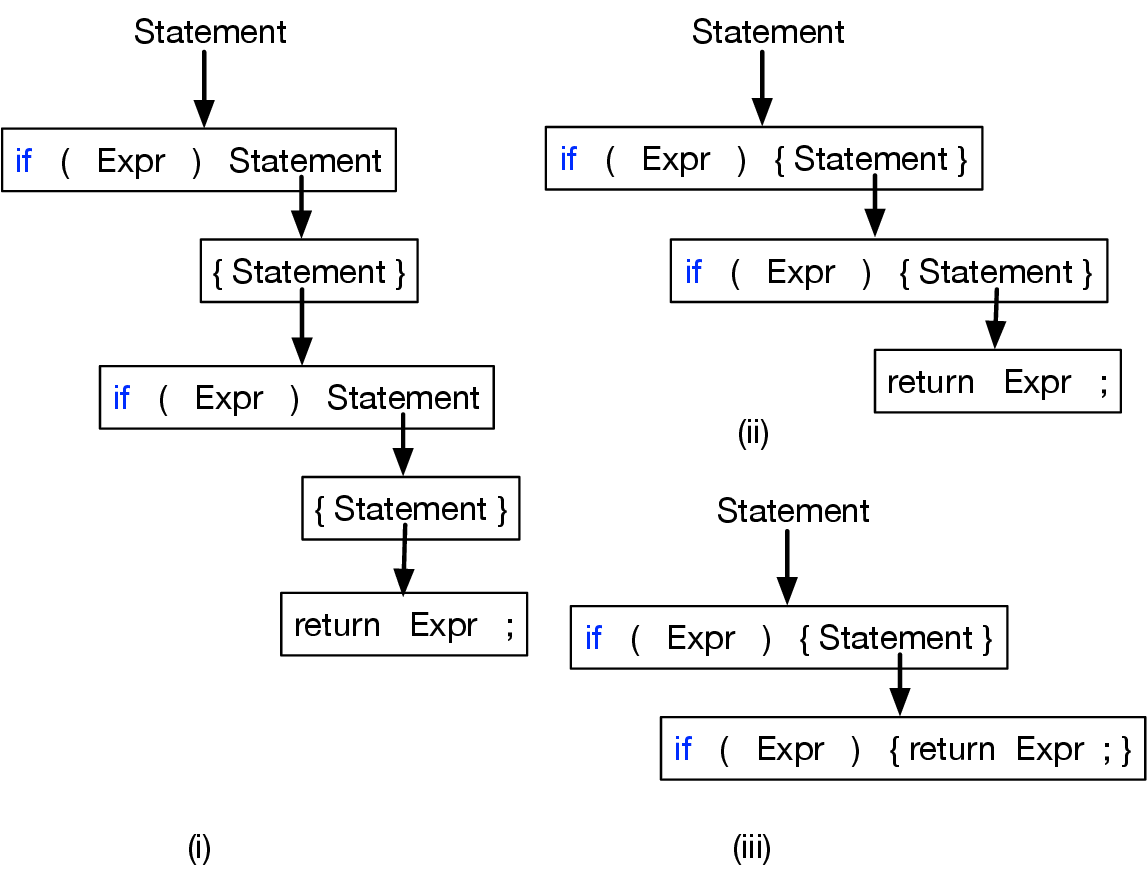}
\caption{(i) Application of the two idioms extracted in Figure \ref{fig:idiom_extraction} on a new training example. (ii) We first perform two applications of the idiom in Figure \ref{fig:idiom_extraction}b, (iii)  followed by an application of the idiom in Figure \ref{fig:idiom_extraction}d. The resulting tree can be represented with just 2 parsing rules instead of 5.}
\label{fig:idiom_application}
\end{figure}

\section{Model Training with Idioms}

Once a set of idioms $I$ is obtained, we next train semantic parsing models to apply these idioms while decoding. We do this by supervising grammar rule generation in the decoder using a compressed set of rules for each example, using the idiom set $I$ (see Algorithm \ref{alg:compress}). More concretely, we first obtain the parse tree $t_i$ (or grammar rule set $p_i$) for each training program $y_i$ under grammar $\mathscr{G}$ (Step 3) and then greedily collapse each depth-2 subtree in $t_i$ corresponding to every idiom in $I$ (Step 5). Once $t_i$ cannot be further collapsed, we translate $t_i$ into production rules $r_i$ based on the collapsed tree, with $|r_i| \leq |p_i|$ (Step 7). 

\begin{algorithm}[h]
\SetKwProg{myproc}{Procedure}{}{end}
  \myproc{Compress($\mathscr{D}, \mathscr{G}, I$)}{
  \KwIn{$\mathscr{D} \rightarrow$ Training Programs}
  \KwIn{$\mathscr{G} \rightarrow$ Parsing Grammar}
  \KwIn{$I \rightarrow$ Idiom set}

  \For{$d \in \mathscr{D}$}{
    $t \gets  $ Parse-Tree$(d, \mathscr{G})$ \\
    \For{$i \in I$}{
        $t \gets $ Collapse-Subtree(t, i)
    }
    Train decoder using Production-Rules$(t)$ \\
  }
}
\caption{Training Example Compression.}
\label{alg:compress}
\end{algorithm}

This process is illustrated in Figure \ref{fig:idiom_application} where we perform two applications of the first idiom from Figure \ref{fig:idiom_extraction} (b), followed by one application of the second idiom from  Figure \ref{fig:idiom_extraction} (d), after which the tree cannot be further compressed using those two idioms. The final tree can be represented using $|r_i| = 2$ rules instead of the original $|p_i| = 5$ rules. The decoder is then trained similar to previous approaches \cite{yin-neubig:2017:Long} using the compressed set of rules. We observe a rule set compression of more than 50\% in our experiments (Section \ref{sec:results}).

\begin{figure}[t]
\includegraphics[width=\linewidth]{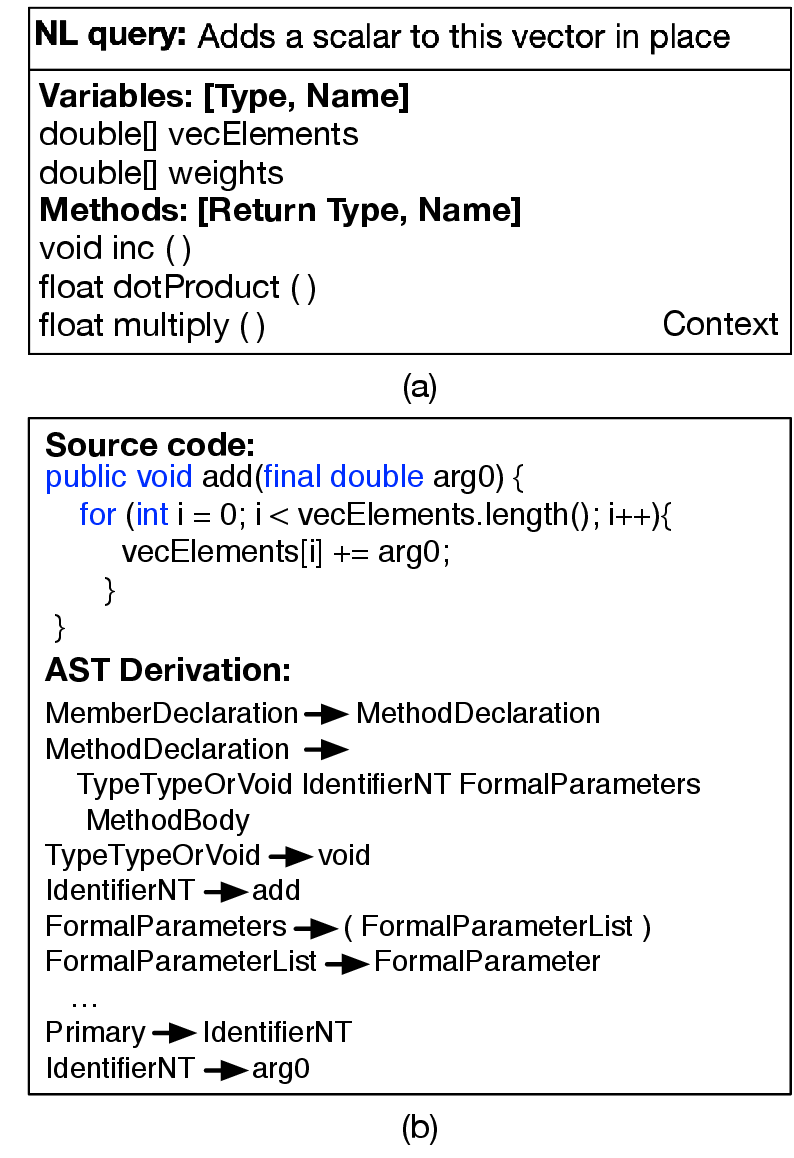}
\caption{Context dependent code generation task of \newcite{iyer2018} that involves mapping an NL query together with a set of context variables and methods (each having an identifier name and data type) (a) into source code, represented as a sequence of grammar rules (b).}
\label{fig:task}
\end{figure}

\section{Experimental Setup}
We apply our approach on 1) the context dependent encoder-decoder model of \newcite{iyer2018} on the CONCODE dataset, where we outperform an improved version of their best model, and 2) the task of mapping NL utterances to SQL queries on the ATIS-SQL dataset \cite{iyer-EtAl:2017:Long} where an idiom-based model using the full training set outperforms the SOTA, also achieving significant gains when using a reduced training set. 

\begin{figure}[t]
\center
\begin{lstlisting}[style=mysql]
%*\textbf{List all flights from Denver to Seattle}*)

SELECT DISTINCT flight_1.flight_id FROM 
 flight f1, airport_service as1, city c1, 
 airport_service as2, city c2 WHERE 
 f1.from_airport = as1.airport_code 
 AND as1.city_code = c1.city_code 
 AND c1.city_name = "Denver" 
 AND f1.to_airport = as2.airport_code 
 AND as2.city_code = c2.city_code 
 AND c2.city_name = "Seattle";

\end{lstlisting}

\caption{Example NL utterance with its corresponding executable SQL query from the ATIS-SQL dataset.}
\label{fig:atis_example}
\end{figure}

\subsection{Context Dependent Semantic Parsing}
\label{sec:semtask}
  The CONCODE task involves mapping an NL query together with a class context comprising a list of variables (with types) and methods (with return types), into the source code of a class member function. Figure \ref{fig:task} (a) shows an example where the context comprises variables and methods (with types) that would normally exist in a class that implements a vector, such as \texttt{vecElements} and \texttt{dotProduct()}. Conditioned on this context, the task involves mapping the NL query \textit{Adds a scalar to this vector in place} into a sequence of parsing rules to generate the source code in Figure \ref{fig:task} (b).

Formally, the task is: Given a NL utterance $q$, a set of context variables $\{v_i\}$ with types $\{t_i\}$, and a set of context methods $\{m_i\}$ with return types $\{r_i\}$, predict a set of parsing rules \{$a_i\}$ of the target program. Their best performing model is a neural encoder-decoder with a context-aware encoder and a decoder that produces a sequence of Java grammar rules.

\paragraph{Baseline Model} We follow the approach of \newcite{iyer2018} with three major modifications in their encoder, which yields improvements in both speed and accuracy (\our). First, in addition to camel-case splitting of identifier tokens, we use byte-pair encoding (BPE) \cite{sennrich-haddow-birch:2016:P16-12} on all NL tokens, identifier names and types and embed all these BPE tokens using a single embedding matrix. Next, we replace their RNN that contextualizes the subtokens of identifiers and types with an average of the subtoken embeddings instead. Finally, we consolidate their three separate RNNs for contextualizing NL, variable names with types, and method names with types, into a single shared RNN, which greatly reduces the number of model parameters. Formally, let $\{q_i\}$ represent the set of BPE tokens of the NL, and $\{t_{ij}\}$, $\{v_{ij}\}, \{r_{ij}\}$ and $\{m_{ij}\}$ represent the $j$th BPE token of the $i$th variable type, variable name, method return type, and method name respectively. First, all these elements are embedded using a BPE token embedding matrix B to give us $\mathbf{q_i}$, $\mathbf{t_{ij}}$, $\mathbf{v_{ij}}$, $\mathbf{r_{ij}}$ and $\mathbf{m_{ij}}$. Using Bi-LSTM $f$, the encoder then computes:
\begin{align}
  h_{1}, \cdots, h_{z}  &= f(\mathbf{q_{1}}, \dots, \mathbf{q_{z}}) \\
  \mathbf{v_i} &= Avg(\mathbf{v_{i1}}, \dots, \mathbf{v_{ij}}) \\ 
  \hspace{30pt} \text{Similarly,} &\text{ compute } \mathbf{m_i, t_i, r_i}\\
  \hat{t_i},\hat{v_i} &= f(\mathbf{t_i}, \mathbf{v_i})  \\
  \hat{r_i},\hat{m_i} &= f(\mathbf{r_i}, \mathbf{m_i}) 
\end{align}

Then, $h_1, \dots, h_z$, and $\hat{t_i},\hat{v_i}, \hat{r_i},\hat{m_i}$ are passed to the attention mechanism in the decoder, exactly as in \newcite{iyer2018}. The decoder remains the same as described in \newcite{iyer2018}, and produces a probability distribution over grammar rules at each time step (full details in Supplementary Materials). This forms our baseline model (\our). 

\paragraph{Idiom Aware Training}
To utilize idioms, we augment this decoder by retrieving the top-K most frequent idioms from the training set (Algorithm \ref{alg:extract}), followed by post-processing the training set by greedily applying these idioms (Algorithm \ref{alg:compress}; we denote this model as \our-K). We evaluate all our models on the CONCODE dataset which was created using Java source files from \url{github.com}. It contains 100K tuples of (NL, code, context) for training, 2K tuples for development, and 2K tuples for testing. We use a BPE vocabulary of 10K tokens (for matrix $B$) and get the best validation set results using the original hyperparameters used by \newcite{iyer2018}. Since idiom aware training is significantly faster than without idioms, it enables us to train on an additional 400K training examples that \newcite{iyer2018} released as part of CONCODE. We report exact match accuracy, corpus level BLEU score (which serves as a measure of partial credit) \cite{papineni2002bleu}, and training time for all these configurations.

\subsection{Semantic Parsing to SQL}

This task involves mapping NL utterances into executable SQL queries. We use the ATIS-SQL dataset \cite{iyer-EtAl:2017:Long} that contains NL questions posed to a flight database, together with their SQL queries and a database with 25 tables to execute them against (Figure \ref{fig:atis_example} shows an example). The dataset is split into 4,379 training, 491 validation, and 448 testing examples following \newcite{kwiatkowski2011lexical}. 

\begin{table}
\centering
\begin{tabular}{lll} 
\toprule
\textbf{Model} &  \textbf{Exact} & \textbf{BLEU}\\
\midrule
Seq2Seq$\dagger$ & 3.2 (2.9) & 23.5 (21.0) \\
Seq2Prod$\dagger$ & 6.7 (5.6) & 21.3 (20.6)  \\ 
\newcite{iyer2018}$\dagger$ & 8.6 (7.1) & 22.1 (21.3)  \\
\midrule
{\our} & 12.5 (9.8) & 24.4 (23.2) \\
{\our} + 200 idioms & 12.2 (9.8) & \textbf{26.6 (24.0)} \\
\bottomrule
\end{tabular}
\caption{Exact Match and BLEU scores for our simplified model (\our) with and without idioms, compared with results from \newcite{iyer2018}$^{\dagger}$ on the test (validation) set of CONCODE. {\our} achieves significantly better EM and BLEU score and reduces training time from 40 hours to 27 hours. Augmenting the decoding process with 200 code idioms further pushes up BLEU and reduces training time to 13 hours.}
\label{tab:results}
\end{table}

The SOTA by \newcite{iyer-EtAl:2017:Long} is a Seq2Seq model with attention and achieves a denotational accuracy of 82.5\% on the test set. Since using our idiom-based approach requires a model that uses grammar-rule based decoding, we use a modified version of the Seq2Prod model described in \newcite{iyer2018} (based on \newcite{yin-neubig:2017:Long}) as a baseline model (Seq2Prod), and augment the decoder with SQL idioms (Seq2Prod-K). 

\begin{table}
\centering
\begin{tabular}{llll} 
\toprule
\textbf{Model} &  \textbf{Exact} & \textbf{BLEU} & \textbf{Training} \\
 &   & & \textbf{Time (h)} \\
\midrule
Seq2Seq$\dagger$ & 2.9 & 21.0 & 12 \\
Seq2Prod$\dagger$ & 5.6 & 20.6 & 36  \\ 
\newcite{iyer2018}$\dagger$ & 7.1 & 21.3 & 40 \\
\midrule
{\our} & 9.8 & 23.2 & 27\\
 + 100 idioms & 9.8 & 24.5 & 15 \\
 + 200 idioms & 9.8 & 24.0 & 13 \\
 + 300 idioms & 9.6 & 23.8 & 12 \\
 + 400 idioms & 9.7 & 23.8 & 11 \\
 + 600 idioms & 9.9 & 22.7 & 11 \\
\bottomrule
\end{tabular}
\caption{Variation in Exact Match, BLEU score, and training time on the validation set of CONCODE with number of idioms used. After top-200 idioms are used, accuracies start to reduce, since using more specialized idioms can hurt model generalization. Training time plateaus after considering top-600 idioms.}
\label{tab:numidioms}
\end{table}

\begin{table}
\centering
\begin{tabular}{llll} 
\toprule
\textbf{Model} &  \textbf{Exact} & \textbf{BLEU} \\
\midrule
1$\times$ Train & 12.0 (9.7) & 26.3 (23.8) \\
2$\times$ Train & 13.0 (10.3) & 28.4 (25.2) \\
3$\times$ Train & 13.3 (10.4)  & 28.6 (26.5) \\
5$\times$ Train & 13.4 (11.0) & 28.9 (26.6) \\
\bottomrule
\end{tabular}
\caption{Exact Match and BLEU scores on the test (validation) set of CONCODE by training \our-400 on the extended training set released by \newcite{iyer2018}. Significant improvements in training speed after incorporating idioms makes training on large amounts of data possible.}
\label{tab:scale}
\end{table}

 Seq2Prod is an encoder-decoder model, where the encoder executes an n-layer bi-LSTM over NL embeddings and passes the final layer LSTM hidden states to an attention mechanism in the decoder. Note that the Seq2Prod encoder described in \newcite{iyer2018} encodes a concatenated sequence of NL and context, but ATIS-SQL instances do not include contextual information. Thus, if $q_i$ represents each lemmatized token of the NL, they are first embedded using a token embedding matrix $B$ to give us $\mathbf{q_i}$. Using Bi-LSTM $f$, the encoder then computes:
\begin{align}
  h_{1}, \cdots, h_{z}  &= f(\mathbf{q_{1}}, \dots, \mathbf{q_{z}})
\end{align}
Then, $h_1, \dots, h_z$ are passed to the attention mechanism in the decoder. 

 The sequential LSTM-decoder uses attention and produces a sequence of grammar rules $\{a_t\}$. The decoder hidden state at time $t$, $s_t$, is computed based on an embedding of the current non-terminal $n_t$ to be expanded, an embedding of the previous production rule $a_{t-1}$, an embedding of the parent production rule, par$(n_t)$, that produced $n_t$, the previous decoder state $s_{t-1}$, and the decoder state of the LSTM cell that produced $n_t$, denoted as $s_{n_t}$.
\begin{align}
  s_t &= \text{LSTM}_f(n_t, a_{t-1}, \text{par}(n_t), s_{t-1}, s_{n_t})
\end{align}
$s_t$ is then used for attention and finally, produces a distribution over grammar rules. 

We make two modifications in this decoder. First, we remove the dependence of $\text{LSTM}_f$ on the parent LSTM cell state $s_{n_t}$. Second, instead of using direct embeddings of rules $a_{t-1}$ and $\text{par}(n_t)$ in LSTM$_f$, we use another Bi-LSTM across the left and right sides of the rule (using separator symbol \texttt{SEP}) and use the final hidden state as inputs to LSTM$_f$ instead. More concretely, if a grammar rule is represented as $A \rightarrow B_1 \dots B_n$, then:
\begin{multline}
\text{Emb} (A \rightarrow B_1 \dots B_n) = \\
\text{LSTM}_g(\{A, \texttt{SEP}, B_1, \dots, B_n\}) \\
s_t = \text{LSTM}_f(n_t, \text{Emb}(a_{t-1}), \text{Emb}(\text{par}(n_t)), \\
s_{t-1})
\end{multline}

This modification can help the LSTM$_f$ cell locate the position of $n_t$ within rules $a_{t-1}$ and $\text{par}(n_t)$, especially for lengthy idiomatic rules. We present a full description of this model with all hyperparameters in the supplementary materials. 

\paragraph{Idiom Aware Training} As before, we augment the set of decoder grammar rules with top-K idioms extracted from ATIS-SQL. To represent SQL queries as grammar rules, we use the python \texttt{sqlparse} package. 

\begin{table}
\centering
\begin{tabular}{ll} 
\toprule
\textbf{Model} &  \textbf{Accuracy}\\
\midrule
\newcite{iyer-EtAl:2017:Long}$\dagger$ & 82.5 \\
Seq2Prod & 79.1  \\ 
\textbf{Seq2Prod + 400 idioms} & \textbf{83.2} \\
\bottomrule
\end{tabular}
\caption{Denotational Accuracy for Seq2Prod with and without idioms, compared with results from \newcite{iyer-EtAl:2017:Long}$^{\dagger}$ on the test set of ATIS using SQL queries. Results averaged over 3 runs.}
\label{tab:atisresults}
\end{table}

\begin{figure}[t]
\includegraphics[width=\linewidth]{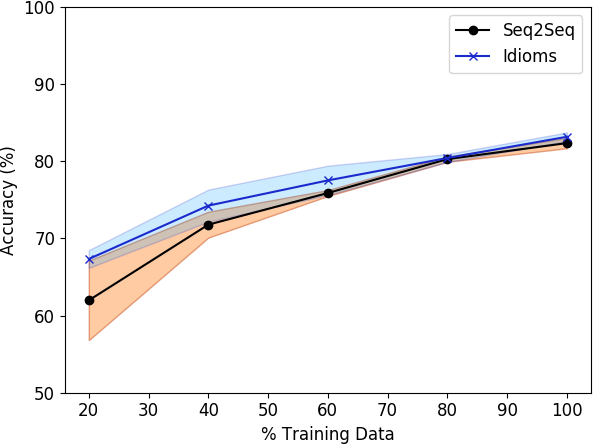}
\caption{Accuracy of the Seq2Seq baseline \cite{iyer-EtAl:2017:Long} compared with Seq2Prod-K, on the test set of ATIS-SQL for varying fractions of training data. Idioms significantly boost accuracy when training data is scarce. (Mean and Std. Dev. computed across 3 runs). 20-40\% use 100 idioms and the rest use 400 idioms.}
\label{fig:learningcurve}
\end{figure}

\begin{figure*}[t]
\includegraphics[width=\textwidth]{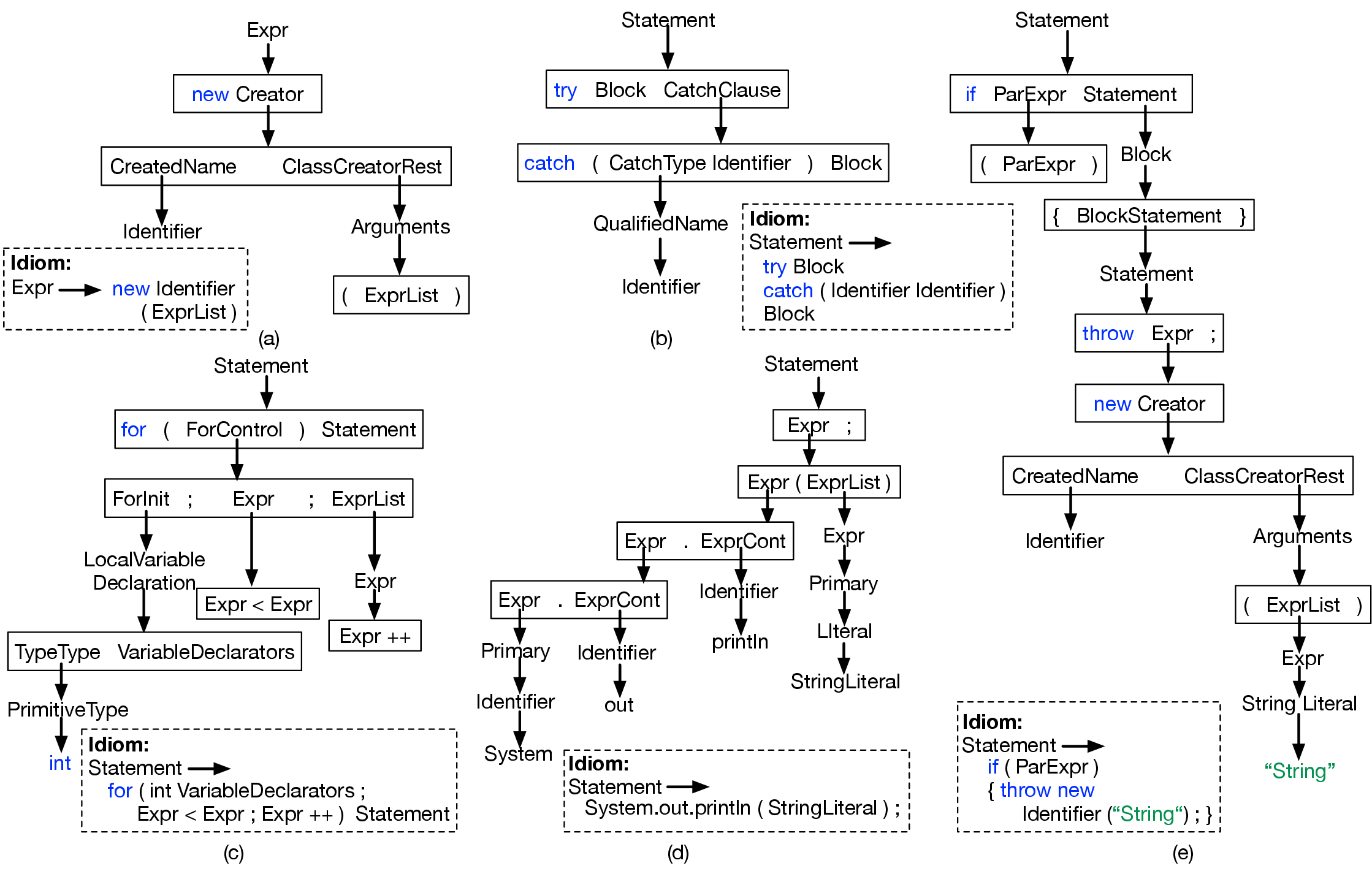}
\caption{Examples of idioms learned from CONCODE. (a)-(c) represent idioms for instantiation of a new object, exception handling and integer-based looping respectively. (d) represents an idiom for applying a very commonly used library method ({\tt System.out.println}). (e) is a combination of various idioms viz. an if-then idiom, an idiom for throwing exceptions and finally, reusing idiom (a) for creating new objects.}
\label{fig:idiom_examples}
\end{figure*}

\section{Results and Discussion}
\label{sec:results}
Table \ref{tab:results} presents exact match and BLEU scores on the original CONCODE train/validation/test split. {\our} yields a large improvement of 3.9 EM and 2.2 BLEU over the best model of \newcite{iyer2018}, while also being significantly faster (27 hours for 30 training epochs as compared to 40 hours). Using a reduced BPE vocabulary makes the model memory efficient, which allows us to use a larger batch size that in turn speeds up training. 
Furthermore, using 200 code idioms further improves BLEU by 2.2\% while maintaining comparable EM accuracy. Using the top-200 idioms results in a target AST compression of more than 50\%, which results in fewer decoder RNN steps being performed. This reduces training time further by more than 50\%, from 27 hours to 13 hours. 

In Table \ref{tab:numidioms}, we illustrate the variations in EM, BLEU and training time with the number of idioms. We find that 200 idioms performs best overall in terms of balancing accuracy and training time. Adding more idioms continues to reduce training time, but accuracy also suffers. Since we permit idioms to contain identifier names to capture frequently used library methods, having too many idioms hurts generalization, especially since the test set is built using repositories disjoint from the training set. Finally, the amount of compression, and therefore the training time, plateaus after the top-600 idioms are incorporated. 

Compared to the model of \newcite{iyer2018}, our significantly reduced training time enables us to train on their extended training set. We run {\our} using 400 idioms (taking advantage of even lower training time) on up to 5 times the amount of data, while making sure that we do not include in training any NL from the validation or the test sets. Since the original set of idioms learned from the original training set are quite general, we directly use them rather than relearn the idioms from scratch. We report EM and BLEU scores for different amounts of training data on the same validation and test sets as CONCODE in Table \ref{tab:scale}. In general, accuracies increase with the amount of data with the best model achieving a BLEU score of 28.9 and EM of 13.4. 

Figure \ref{fig:idiom_examples} shows example idioms extracted from CONCODE: (a) is an idiom to construct a new object with arguments, (b) represents a {\tt  try}-{\tt catch} block, and, (c) is an integer-based {\tt for} loop. In (e), we show how small idioms are combined to form larger ones; it combines an {\tt if}-{\tt then} idiom with a {\tt throw}-{\tt exception} idiom, which throws an object instantiated using idiom (a). The decoder also learns idioms to directly generate common library methods such as \texttt{System.out.println( StringLiteral )} in one decoding step (d).

For the NL to SQL task, we report denotational accuracy in Table \ref{tab:atisresults}. We observe that Seq2Prod underperforms the Seq2Seq model of \newcite{iyer-EtAl:2017:Long}, most likely because a SQL query parse is much longer than the original query. This is remedied by using top-400 idioms, which compresses the decoded sequence size, marginally outperforming the SOTA (83.2\%). \newcite{finegan2018improving} observed that the SQL structures in ATIS-SQL are repeated numerous times in both train and test sets, thus facilitating Seq2seq models to memorize these structures without explicit idiom supervision. To test a scenario with limited repetition of structures, we compare Seq2Seq with Seq2Prod-K for limited training data (increments of 20\%) and observe that (Figure \ref{fig:learningcurve}) idioms are additionally helpful with lesser training data, consistent with our intuition.

\section{Conclusions}

We presented a general approach to make semantic parsers aware of target idiomatic structures, by first identifying frequently used idioms, followed by providing models with supervision to apply these idioms. We demonstrated this approach on the task of context dependent code generation where we achieved a new SOTA in EM accuracy and BLEU score. We also found that decoding using idioms significantly reduces training time and allows us to train on significantly larger datasets. Finally, our approach also outperformed the SOTA for a semantic parsing to SQL task on ATIS-SQL, with significant improvements under a limited training data regime. 

\section*{Acknowledgements}

University of Washington's efforts in this research was supported in part by DARPA (FA8750-16-2-0032), the ARO (ARO-W911NF-16-1-0121), the NSF (IIS-1252835, IIS-1562364, IIS-1546083, IIS-1651489, OAC-1739419), the DOE (DE-SC0016260), the Intel-NSF CAPA center, and gifts from Adobe, Amazon, Google, Huawei, and NVIDIA. The authors thank the anonymous reviewers for their helpful comments.

\bibliography{emnlp2019}
\bibliographystyle{acl_natbib}

\appendix
\section*{Supplementary Materials}

\section{{\our} for CONCODE}

{\our} is similar to the best performing encoder-decoder model of \newcite{iyer2018} on the CONCODE dataset, with three major modifications in their encoder, which yields improvements in speed and accuracy. First, in addition to camel-case splitting of identifier tokens, we use byte-pair encoding (BPE) \cite{sennrich-haddow-birch:2016:P16-12} on all NL tokens, identifier names and types and embed these BPE tokens using a single embedding matrix. Next, we replace their RNN that contextualizes the subtokens of identifiers and types with an average of the subtoken embeddings instead. Finally, we consolidate their three separate RNNs for contextualizing NL, variable names with types, and method names with types, into a single shared RNN, which greatly cuts down model parameters. We present the full model here for convenience. Since the decoder is unmodified, large portions of this section are borrowed from \newcite{iyer2018}.

Formally, let $\{q_i\}$ represent the set of BPE tokens of the NL, and $\{t_{ij}\}$, $\{v_{ij}\}, \{r_{ij}\}$ and $\{m_{ij}\}$ represent the $j$th BPE token of the $i$th variable type, variable name, method return type, and method name respectively.

\paragraph{Encoder} The encoder computes contextual representations of the NL and each component in the context.  First, all the elements defined above are embedded using a BPE token embedding matrix $B$ to give us $\mathbf{q_i}$, $\mathbf{t_{ij}}$, $\mathbf{v_{ij}}$, $\mathbf{r_{ij}}$ and $\mathbf{m_{ij}}$. Using Bi-LSTM $f$, the encoder then computes:

\begin{align}
  h_{1}, \cdots, h_{z}  &= f(\mathbf{q_{1}}, \dots, \mathbf{q_{z}}) \\
  \mathbf{v_i} &= Avg(\mathbf{v_{i1}}, \dots, \mathbf{v_{ij}}) \\ 
  \hspace{30pt} \text{Similarly,} &\text{ compute } \mathbf{m_i, t_i, r_i}\\
  \hat{t_i},\hat{v_i} &= f(\mathbf{t_i}, \mathbf{v_i})  \\
  \hat{r_i},\hat{m_i} &= f(\mathbf{r_i}, \mathbf{m_i}) 
\end{align}

Then, $h_1, \dots, h_z$, and $\hat{t_i},\hat{v_i}, \hat{r_i},\hat{m_i}$ are passed to the attention mechanism in the decoder. 

\paragraph{Decoder} The decoder is a sequential LSTM based model that produces a sequence of grammar rules ($a_t$ at step $t$), which can later be put together to form a source code snippet. At each time step $t$, the decoder expands a non-terminal that was produced earlier, by choosing a valid right hand side rule for that non-terminal. The first non-terminal (at step 1) is \textit{MemberDeclaration} and subsequently, every non-terminal is expanded in a depth first left to right fashion, similar to \newcite{yin-neubig:2017:Long}. The probability of a source code snippet is decomposed as a product of the conditional probability of generating each step in the sequence of rules conditioned on the previously generated rules. 

More specifically, the decoder is an LSTM-based RNN with hidden dimension size $H$,  that produces a context vector $c_t$ at each time step, which is used to compute a distribution over next actions. 
\begin{align} \label{eq:1}
  p(a_t|a_{<t}) &\propto exp(W^{n_t} c_t)
\end{align}
Here, $W^{n_t}$ is a $|n_t| \times H$ matrix, where $|n_t|$ is the total number of unique grammar rules that $n_t$ can be expanded to. The context vector $c_t$ is computed using the hidden state $s_t$ of an n-layer decoder LSTM cell and attention vectors over the NL and the context ($z_t$ and $e_t$), as described below. 

\paragraph{Decoder LSTM} The decoder uses an n-layer LSTM whose hidden state $s_t$ is computed based on the current non-terminal $n_t$ to be expanded, the previous production rule $a_{t-1}$, the parent production rule, par$(n_t)$, that produced $n_t$, the previous decoder LSTM state $s_{t-1}$, and the decoder state of the LSTM cell that produced $n_t$, denoted as $s_{n_t}$. \begin{align}
  s_t &= \text{LSTM}(n_t, a_{t-1}, \text{par}(n_t), s_{t-1}, s_{n_t})
\end{align}
We use an embedding matrix $N$ to embed $n_t$ and matrix $A$ to embed $a_{t-1}$ and $\text{par}(n_t)$. If $a_{t-1}$ is a rule that generates a terminal node that represents an identifier or literal, it is represented using a special rule \textit{IdentifierOrLiteral} to collapse all these rules into a single previous rule. 

\paragraph{Two-step Attention} At time step $t$, the decoder first attends to every token in the NL representation, $h_i$, using the current decoder state, $s_t$, to compute a set of attention weights $\alpha_t$, which are used to combine $h_i$ into an NL context vector $z_t$. We use a general attention mechanism~\cite{luong-pham-manning:2015:EMNLP}, extended to perform multiple steps.
\begin{align*}
\alpha_{t, i} &= \frac{\text{exp}(s_t^\text{T} \mathbf{F} h_i)}{\sum_{i} \text{exp}(s_t^\text{T} \mathbf{F} h_i)} \\
z_t &=  \sum_{i} \alpha_{t, i} h_i
\end{align*}
The context vector $z_t$ is used to attend over every type (return type) and variable (method) name in the environment, to produce attention weights $\beta_t$ that are used to combine the entire context \mbox{$x=[t:v:r:m]$} into an environment context vector $e_t$.\footnote{``:'' denotes concatenation.}
\begin{align*}
\beta_{t, j} &= \frac{\text{exp}(z_t^\text{T} \mathbf{G} x_j)}{\sum_{j} \text{exp}(z_t^\text{T} \mathbf{G} x_j)} \\
e_t &=  \sum_{j} \beta_{t, j} x_j
\end{align*}
Finally, $c_t$ is computed using the decoder state and both context vectors $z_t$ and $e_t$:
\begin{align*}
  c_t = tanh(\hat{W}[s_t:z_t:e_t])
\end{align*}

\paragraph{Supervised Copy Mechanism} Since the class environment at test time can belong to previously unseen new domains, our model needs to learn to copy variables and methods into the output. We use the copying technique of \newcite{gu-EtAl:2016:P16-1} to compute a copy probability at every time step $t$ using learned vector $b$ of dimensionality $H$.
\begin{align*}
   \text{copy}(t) = \sigma(b^T c_t)
\end{align*}
Since we only require named identifiers or user defined types to be copied, both of which are produced by production rules with $n_t$ as \textit{IdentifierNT}, we make use of this copy mechanism only in this case. Identifiers can be generated by directly generating derivation rules (see equation \ref{eq:1}), or by copying from the environment. The probability of copying an environment token $x_j$, is set to be the attention weights $\beta_{t,j}$ computed earlier, which attends exactly on the environment types and names which we wish to be able to copy. The copying process is supervised by preprocessing the grammar rules to recognize identifiers that can be copied from the environment, and both the generation and the copy probabilities are weighted by $1 - \text{copy}(t)$ and $\text{copy}(t)$ respectively.

\paragraph{Hyperparameters and Inference} We use an embedding size $H$ of 1024 for identifiers and types. All LSTM cells use 2-layers and a hidden dimensionality of 1024 (512 on each direction for BiLSTMs). We use an embedding size of $512$ for encoding non-terminals and rules in the decoder. We use dropout with $p=0.5$ in between LSTM layers and at the output of the decoder over $c_t$. We train our model for $30$ epochs using mini-batch gradient descent with a batch size of $40$, and we use Adam \cite{kingma2014adam} with an initial learning rate of $0.001$ for optimization. We decay our learning rate by $80\%$ based on performance on the development set after every epoch. We use beam search with a beam size of 5 for decoding the sequence of grammar rules at test time. 

\section{Seq2Prod for ATIS-SQL}

Our Seq2Prod is similar to the Seq2Prod model of \newcite{iyer2018}, with a modification in the inputs to the decoder LSTM. We describe the entire model here for convenience. Large portions of this section are borrowed from \newcite{iyer2018}.

\paragraph{Encoder} The encoder of Seq2Prod computes contextual representations of the NL query. Note that unlike the previous model for CONCODE, this model does not need to encode context. If $q_i$ represents each lemmatized token of the NL, they are first embedded using a token embedding matrix $B$ to give us $\mathbf{q_i}$. Using Bi-LSTM $f$, the encoder then computes:
\begin{align}
  h_{1}, \cdots, h_{z}  &= f(\mathbf{q_{1}}, \dots, \mathbf{q_{z}})
\end{align}
Then, $h_1, \dots, h_z$ are passed to the attention mechanism in the decoder. 

\paragraph{Decoder} Similar to the {\our} model described above, the decoder is a sequential LSTM based model that produces a sequence of grammar rules ($a_t$ at step $t$), which can later be put together to form a source code snippet. More specifically, the decoder is an LSTM-based RNN with hidden dimension size $H$,  that produces a context vector $c_t$ at each time step, which is used to compute a distribution over next actions. 
\begin{align} 
  p(a_t|a_{<t}) &\propto exp(W^{n_t} c_t)
\end{align}
Here, $W^{n_t}$ is a $|n_t| \times H$ matrix, where $|n_t|$ is the total number of unique grammar rules that $n_t$ can be expanded to. The context vector $c_t$ is computed using the hidden state $s_t$ of an n-layer decoder LSTM cell and attention vectors over the NL $z_t$, as described below. 

\paragraph{Decoder LSTM} The decoder uses an n-layer LSTM (LSTM$_f$) whose hidden state at time $t$, $s_t$, is computed based on an embedding of the current non-terminal $n_t$ to be expanded, a contextuatized embedding of the previous production rule $a_{t-1}$, a contextualized embedding of the parent production rule, par$(n_t)$, that produced $n_t$, and the previous decoder LSTM state $s_{t-1}$. Note that unlike the previous {\our} model, we do not use the parent LSTM state as it does not provide any improved performance.

We use an embedding matrix $N$ to embed $n_t$. To compute the contextualized embeddings of $a_{t-1}$ and $\text{par}(n_t)$ in LSTM$_f$, we use another single layer Bi-LSTM (LSTM$_g$) across the left and right sides of the rule (using separator symbol \texttt{SEP}) and use the final hidden state as inputs to LSTM$_f$ instead. More concretely, if a grammar rule is represented as $A \rightarrow B_1 \dots B_n$, then:
\begin{multline}
\text{Emb} (A \rightarrow B_1 \dots B_n) = \\
\text{LSTM}_g(\{A, \texttt{SEP}, B_1, \dots, B_n\})
\end{multline} 
\vspace{-25pt}
\begin{multline}
s_t = \text{LSTM}_f(n_t, \text{Emb}(a_{t-1}), \text{Emb}(\text{par}(n_t)), \\
s_{t-1})
\end{multline}

The contextualization of these rule embeddings is the primary difference between our model and \newcite{iyer2018}. This modification can help the LSTM$_f$ cell locate the position of $n_t$ within rules $a_{t-1}$ and $\text{par}(n_t)$, especially, for lengthy idiomatic rules. 

$s_t$ is then used for attention and finally, produces a distribution over grammar rules. 

\paragraph{Single-Step Attention} At time step $t$, the decoder attends to every token in the NL representation, $h_i$, using the current decoder state, $s_t$, to compute a set of attention weights $\alpha_t$, which are used to combine $h_i$ into an NL context vector $z_t$. We use the general attention  mechanism of \newcite{luong-pham-manning:2015:EMNLP}.
\begin{align*}
\alpha_{t, i} &= \frac{\text{exp}(s_t^\text{T} \mathbf{F} h_i)}{\sum_{i} \text{exp}(s_t^\text{T} \mathbf{F} h_i)} \\
z_t &=  \sum_{i} \alpha_{t, i} h_i
\end{align*}

Finally, $c_t$ is computed using the decoder state and context vector $z_t$:
\begin{align*}
  c_t = tanh(\hat{W}[s_t:z_t])
\end{align*}

\paragraph{Supervised Copy Mechanism} The supervised copy mechanism is exactly the same as described for the previous model (\our), using context vector $c_t$. 

\paragraph{Hyperparameters} We use an embedding size $H$ of 1024 for NL query tokens. Both the encoder and decoder LSTM cells use 2-layers and a hidden dimensionality of 1024 (512 on each direction for BiLSTMs). We use an embedding size of $512$ for encoding non-terminals and a hidden size of $256$ for the contextualized rules (for LSTM$_g$) in the decoder. We use dropout with $p=0.5$ in between LSTM layers and at the output of the decoder over $c_t$. We train our model for $60$ epochs using mini-batch gradient descent with a batch size of $40$, and we use Adam \cite{kingma2014adam} with an initial learning rate of $0.001$ for optimization. We decay our learning rate by $80\%$ based on performance on the development set after every epoch. We use beam search with a beam size of 5 for decoding the sequence of grammar rules at test time.  

\end{document}